\documentclass[sigconf]{acmart}

\usepackage{booktabs} 

\usepackage{enumerate}

\copyrightyear{2017}
\acmYear{2017}
\setcopyright{acmlicensed}
\acmConference{GECCO '17 Companion}{July 15-19, 2017}{Berlin, Germany}\acmPrice{15.00}\acmDOI{http://dx.doi.org/10.1145/3067695.3082529} \acmISBN{978-1-4503-4939-0/17/07}

\DeclareMathOperator*{\argmin}{argmin}
\DeclareMathOperator*{\argmax}{argmax}

\begin{document}
\title{Genealogical Distance as a Diversity Estimate\\in Evolutionary Algorithms}

\author{Thomas Gabor}
\affiliation{%
  \institution{LMU Munich}
}
\email{thomas.gabor@ifi.lmu.de}

\author{Lenz Belzner}
\affiliation{%
  \institution{LMU Munich}
}
\email{belzner@ifi.lmu.de}

\renewcommand{\shortauthors}{Gabor and Belzner}

\begin{abstract}
The evolutionary edit distance between two individuals in a population, i.e., the amount of applications of any genetic operator it would take the evolutionary process to generate one individual starting from the other, seems like a promising estimate for the diversity between said individuals. We introduce genealogical diversity, i.e., estimating two individuals' degree of relatedness by analyzing large, unused parts of their genome, as a computationally efficient method to approximate that measure for diversity.
\end{abstract}

%
%




\maketitle

\section{Introduction}

Diversity has been a central point of research in the area of evolutionary algorithms. It is a well-known fact that maintaining a certain level of diversity aids the evolutionary process in preventing \emph{premature convergence}, i.e., the phenomenon that the population focuses too quickly on a local optimum at hand and never reaches more fruitful areas of the fitness landscape \cite{ursem2002diversity, eiben2003introduction, squillero2016divergence}. Diversity thus plays a key role in adjusting the \emph{exploration-exploitation trade-off} found in any kind of metaheuristic search algorithm.  

We encountered this problem from an industry point of view when designing learning components for a system that needs to guarantee certain levels of quality despite being subjected to the probabilistic nature of its physical environment and probabilistic behavior of its machine learning parts \cite{belzner2016software}. Of course, any general solution for this kind of challenge is yet to be found. However, we believe that the engineering of (hopefully) reliable learning components can be supported by exposing all handles that search algorithms offer to the engineer at site. For scenarios like this one, we consider it most helpful to employ approaches that allow the engineer to actively control properties like diversity of the evolutionary search process instead of just observing diversity and adjust it indirectly via other parameters (like, e.g., the mutation rate).

Among the copious amount of different techniques to introduce diversity-awareness to evolutionary algorithms, many do not immediately make the job of adjusting a given evolutionary algorithm easier but instead require additional engineering effort: For example, one may need to define a distance metric specifically for the search domain at hand or adjust lots of hyperparameters in island or niching models. We thus attempt to define a more domain-independent and almost parameter-free measurement for diversity by utilizing the genetic operators already defined within any given evolutionary process.

We discuss related work in the following Section~\ref{related}. We then explain the target metric called ``evolutionary edit distance'' in Section~\ref{edit}. Section~\ref{tree} continues by introducing the notion of ``genealogical diversity'' as means to approximate that concept. We improve this approach in Section~\ref{trash} by using a much simpler and computationally more efficient data structure. To support our ideas, we describe a basic evaluation scenario in which we have applied both approaches in Section~\ref{experiment}. We end with a short conclusion in Section~\ref{conclusion}.

\section{Related Work}
\label{related}

The importance of diversity for evolutionary algorithms is discussed throughout the body of literature on evolutionary computing ranging from entry level \cite{eiben2003introduction, kruse2016computational} to specialized papers \cite{ursem2002diversity, segura2016using}. In many cases, authors refer to diversity as a measure of the evolutionary algorithm's performance and try to configure the hyperparameters of the evolutionary algorithm as to achieve an optimal trade-off between exploration and exploration for the scenario at hand \cite{tomasini2005spatially}. This measure can then be used to interact with the evolutionary process by adjusting its parameters \cite{ursem2002diversity} and/or actively altering the current population, for example via episodes of ``hypermutation'' \cite{grefenstette1992genetic} or migration of individuals from other (sub-)populations \cite{tomasini2005spatially, li2017subpopulation}. On top of that, there exist a few approaches that include diversity into the evolutionary algorithm's objective function allowing us to use evolution's optimization abilities to explicitly achieve higher diversity of solutions \cite{brameier2002explicit}.

An extensive overview of current approaches to increase diversity in evolutionary algorithms is provided in \cite{squillero2016divergence}, which also defines a helpful taxonomy of said approaches. Whenever a diversity objective can be quantified, it can be used to build a classic multi-objective optimization problem and to apply the vast amount of techniques developed to solve these kinds of problems using evolutionary algorithms \cite{horn1994niched, laumanns2002combining, konak2006multi, segura2016using}.

The authors of \cite{wineberg2003underlying} address the very important issue of how to efficiently compute diversity estimates requiring to compare every individual of a population to every other. They develop an approach to reduce the complexity of said computation to linear time. However, it might still be interesting to analyze how well certain metrics scale even beyond that, as for example in the present paper we chose to sample the test set for diversity from the population to further save computation time.

\section{Evolutionary Edit Distance}
\label{edit}

As described in the previous Section, there exists a vast amount of approaches to compute a population's diversity (and an individual's diversity with respect to that population). We found, among other things, that from an engineering point of view, many (if not most) of these approaches require the designer of the evolutionary algorithm to adjust certain functions or parameters based on the problem domain \cite{gabor2017icac}. This gave rise to the idea of using the genetic operators already programmed for the problem domain to define a domain-independent notion of diversity.

The concept this line of thought is based on could be called \emph{evolutionary edit distance}: Given two individuals $x_1$ and $x_2$ we want to estimate how many applications of a genetic operator it would take to turn one of these individuals into the other.\footnote{Because of the probabilistic nature of evolutionary algorithms, the evolutionary edit distance would actually be a distribution over integers. If a scalar value is needed, we could then compute the \emph{expected evolutionary edit distance}.} First, we start off by defining a lower bound on the number of operator applications, i.e., the \emph{minimal evolutionary edit distance}.

We can assume that a given evolutionary process provides the genetic operator $o: \mathcal{D}^* \to \mathcal{D}$ where $\mathcal{D}$ is the problem domain in which our individuals live and $\mathcal{D}^*$ is a list of arbitrary many elements of $\mathcal{D}$. Most evolutionary algorithms define exactly two instances of genetic operators called mutation $m: \mathcal{D} \to \mathcal{D}$ and recombination $c: \mathcal{D} \times \mathcal{D} \to \mathcal{D}$, but we describe the more general case for now. However, in the general case genetic operators perform in a probabilistic manner, meaning that their exact results depend on chance. We describe this behavior mathematically by adding an index $s \in \mathcal{S}$ to $o$ which represents the seed of a pseudo-random number generator (using seeds of type $\mathcal{S}$). Then, we can define the minimal evolutionary edit distance $\textit{mdist}: \mathcal{D} \times \mathcal{D} \to \mathbb{N}$ as follows:

$$\textit{mdist}(x_1, x_2) = \begin{cases}
    0 & \textit{if } x_1 = x_2\\
    \min\limits_{s \in \mathcal{S}} 1 + \textit{mdist}(o_s(x_1), x_2) & \textit{otherwise}
\end{cases}$$

Note that as long as we assume the genetic operator $o$ to be symmetric (which they usually are), $\textit{mdist}$ is symmetric as well.

The minimal edit distance is not an accurate estimate of the actual effort it would take the evolutionary process to turn $x_1$ into $x_2$ since the required indexed instances $o_s$ of the genetic operator $o$ may be arbitrarily unlikely to occur in the process. Instead, we want to estimate the expected amount of applications of $o$ given a realistic occurrence of instances of the genetic operator. Sadly, the complexity of this problem is equal to running an evolutionary algorithm optimizing its individuals to look like $x_2$ and thus potentially equally expensive regarding computational effort as the evolutionary process we are trying to augment with diversity.

However, if we want to use $\textit{mdist}$ to compute the diversity of individuals for a given evolutionary process, we never want to compare arbitrary solution candidates $x_1, x_2 \in \mathcal{D}$ but will only ever compare individuals within the current population $P \subseteq \mathcal{D}$ or at most individuals from the set $X$ with $P \subseteq X \subseteq \mathcal{D}$, which is the set of all individuals ever generated by the evolutionary process. Each of those individuals has been generated through the repetitive application of the genetic operators already and so we have a set of concrete instances of $o$ instead of having to reason about all $o_s$ that \emph{could} be used by the evolutionary process. We write the set of all actually generated instances of $o$ as $\mathcal{O} = \{(x_0, o_{s_0}, x'_0), (x_1, o_{s_1}, x'_1)..., (x_k, o_{s_k}, x'_k)\}$ where $k+1$ is the total amount of evolutionary operations performed and for all $(x_i, o_{s_i}, x'_i) \in \mathcal{O}$ the evolutionary process actually constructed $x'_i \in X$ by computing $o_{s_i}(x_i)$.

We can thus define the \emph{factual} evolutionary edit distance $\textit{edist}: X \times X \to \mathbb{N}$ as the total amount of operations it actually took to turn $x$ into $x'$:

$$\textit{edist}(x_1, x_2) = \begin{cases}
    0 & \textit{if } x_1 = x_2\\
    1 & \textit{if } \exists s \in \mathcal{S}: (x_1, o_s, x_2) \in \mathcal{O}\\
    1 & \textit{if } \exists s \in \mathcal{S}: (x_2, o_s, x_1) \in \mathcal{O}\\
    edist'(x_1, x_2)  & \textit{otherwise}
\end{cases}$$

\begin{align*}
\textit{edist}'(x_1, x_2) = & \min\limits_{x \in X} \;\;  1 \; +\\
& \;\;\;\; \min\limits_{s \in \mathcal{S}, (x, o_s, x_1) \in \mathcal{O}} \;\; \textit{edist}(o_s(x), x_1) \; + \\ & \;\;\;\; \min\limits_{s \in \mathcal{S}, (x, o_s, x_2) \in \mathcal{O}} \;\; \textit{edist}(o_s(x), x_2))
\end{align*}

Note that $\textit{edist}$ can only be defined this way when we assume that its parameters $x_1$ and $x_2$ have actually been generated through the application of genetic operators from a single base individual only. This is an unrealistic assumption: Completely unrelated individuals can be generated during evolution. Furthermore, defining the evolutionary edit distance this way requires multiple iterations through the whole set of $X$ since we neglect many restrictions present in most genetic operators $o$.

\section{Paths in the Genealogical Tree}
\label{tree}

In the context of evolutionary processes it seems natural to think of individuals as forming genealogical relationships between each other. These relations correspond to the genetic operators applied to an individual $x$ to create the individual $x'$. Connecting all individuals (of all generations of the evolutionary process) to their respective children yields a directed, acyclic and usually non-connected graph. Starting from a single individual $x$, recursively traversing all incoming edges in reverse direction yields a connected subgraph containing all of $x$'s ancestors. We call this graph the genealogical tree of $x$.

Formally, we write $\mathfrak{G}(x) = (\mathfrak{V}_x, \mathfrak{E}_x)$ for the genealogical tree of $x$ consisting of vertices $\mathfrak{V}_x$ and edges $\mathfrak{E}_x$. For an evolutionary process producing (over all generations) the set of individuals $X$, it holds for all $x_1, x_2 \in X$ that $(x_1, x_2) \in \mathfrak{E}_{x_2}$ iff $x_2$ is the result of a variation of $x_1$. If we consider an evolutionary process with two-parent recombination as its only variation operator, our notion of a genealogical tree is exactly the same as in human (or animal) genealogy.

However, most evolutionary algorithms also feature a mutation operator that works independently from recombination. For the genealogical tree, we treat it like a one-parent recombination in that we consider a mutated individual an ancestor of the original one. This approach does not reflect the fact that a single mutation usually has a much smaller impact on the genome of an individual than recombination has. We tackle this issue in Section~\ref{trash}.

Given these graphs, we can then trivially define the \emph{ancestral distance} from an individual $x_1 \in X$ to another individual $x_2 \in X$ as follows:

$$\textit{adist}(x_1, x_2) = \begin{cases}
    \infty & \textit{if } x_1 \notin \mathfrak{V}_{x_2}\\
    0 & \textit{if } x_1 = x_2\\
    \min\limits_{x \in X, (x, x_2) \in \mathfrak{E}_{x_2}} 1 + \textit{adist}(x_1, x) & \textit{otherwise}
\end{cases}$$

Note that $\textit{adist}$ as defined here is still not symmetric, i.e., it returns the amount of variation steps it took to get from $x_1$ to $x_2$, which is a finite number iff $x_1$ is an ancestor of $x_2$. This also usually means that if $\textit{adist}(x_1, x_2)$ is finite, $\textit{adist}(x_2, x_1) = \infty$.

Given two individuals $x_1$ and $x_2$, we can use these definitions to compute their \emph{latest common ancestor} $L(x_1, x_2)$, i.e., the individual with the closest relationship to either $x_1$ or $x_2$ that appears in the respective other individual's genealogical tree. Formally, if a (latest) common ancestor exists it is given via:

$$L(x_1, x_2) = \argmin\limits_{x \in \mathfrak{V}_{x_1} \cap \mathfrak{V}_{x_2}} \min(\textit{adist}(x, x_1), \textit{adist}(x, x_2))$$

Note that $L$ is symmetric, so $L(x_1, x_2) = L(x_2, x_1)$. For our definition of genealogical distance we consider the ancestral distance from the latest common ancestor to the given individuals. However, we also want to normalize the distance values with respect to the maximally achievable distance for a certain individual's age. The main benefit here is that when normalizing genealogical distance on a scale of $[0; 1]$, e.g., we can assign a finite distance to two completely unrelated individuals. For this reason we define a function $E$ which computes the earliest ancestor of a given individual:

$$E(x) = \argmax\limits_{x' \in \mathfrak{V}_x} \textit{adist}(x', x)$$

Note that for all $x' \in \mathfrak{V}_x$ it holds that $\textit{adist}(x', x)$ is finite. We can now use the ancestral distance to an individual's earliest ancestor to normalize the distance to the latest common ancestor with respect to the age of the evolutionary process. Note that if $x_1$ and $x_2$ share no common ancestor, we set $\textit{gdist}(x_1, x_2) = 1$ and otherwise:

$$\textit{gdist}(x_1, x_2)\!=\!\frac{\min(\textit{adist}(L(x_1, x_2), x_1), \textit{adist}(L(x_1, x_2), x_2))}{\max(\textit{adist}(E(x_1, x_2), x_1), \textit{adist}(E(x_1, x_2), x_2))}$$

This genealogical distance function $\textit{gdist}$ then describes for two individuals $x_1, x_2$ how close their latest common ancestor is in comparison to their combined ``evolutionary age'', i.e., the total amount of variation operations they went through.

Following up from the previous Section, we claim that this genealogical distance correlates to the factual evolutionary edit distance between two individuals. It is not an exact depiction, though, because for cousins, e.g., we choose the minimum distance to their common ancestor instead of adding both paths through which they originated from their ancestor. Our reason for doing so is that we want to treat the comparison of cousins to cousins and of parents to children the same way, but the ancestral distance from child to parent is $\infty$. In the end, we are not interested in the exact values but only in the comparison of various degrees of relatedness, which is why lowering the overall numbers using $\min$ instead of summation seems reasonable.

In effect, the metric of $\textit{gdist}$ still appears to be needlessly exact for the application purpose inside the highly stochastic nature of an evolutionary algorithm. And while a lot of algorithmic optimizations and caching of ancestry values can help to cut down the computation time of the employed metric, comparing two individuals still takes linear time with respect to the node count of their ancestral trees, which in turn is likely to grow over time of the evolutionary process. We tackle these issues by introducing a faster and more heuristic approach in the following Section.

\section{Estimating Genealogical Distance on the Genome}
\label{trash}

At first, it seems impossible or at east overly difficult to estimate the genealogical distance (or for that matter, the evolutionary edit distance) of two individuals without knowing about their ancestry inside the evolutionary process. However, life sciences are facing that problem per usual and have found a way to estimate the relationship between two different genomes rather accurately. They do so by computing the similarity in genetic material between two given genomes.

To most artificial evolutionary processes, this approach is not directly applicable for a few reasons:

\begin{enumerate}[(i)]
    \item Most evolutionary algorithms use genomes that are much smaller than that of living beings. Thus, it is much harder to derive statistical similarity estimates and the analysis is much more prone to be influenced by sampling error.
    \item In many cases, the genomes used are not homogeneous but include various fields of different data types. Comparing similarity between different types of data requires a rather complex combined similarity metric.
    \item The way genomes are usually structured in evolutionary algorithms means that most to all parts of the genome are subject to selection pressure reducing the variety found between different genomes.
\end{enumerate}

The last point may seem odd because, obviously, genomes found in nature are subject to selection pressure as well. However, biology has found that, in fact, most parts of the human genome are not expressed at all when building the phenotype (i.e., a human body) \cite{mills2007transposable} and are thus not directly subjected to selection pressure.

We can, however, mitigate these problems making a rather simple addition to an arbitrary evolutionary algorithm: \emph{Add more genes}. As these additional genes do not carry any meaning for the solution candidate encoded by the genome, they are not subjected to selection pressure (iii). We can choose any data type we want for them, so we can adhere to a type that allows for an easy comparison between individuals (ii). And finally, we can choose a comparatively large size for these genes so that they allow for a subtle comparison (i). For the lack of a better name, we call these additional genes by the name \emph{trash genes} to emphasize that they do not directly contribute to the individual's fitness.

For our experiments thus far, we have chosen a simple bit vector of a fixed length $\tau$ to encode the added trash genes. Choosing $\tau$ too small ($2^\tau < n$ where $n$ is the population size) can obviously be detrimental to the distance estimate, but choosing very large $\tau$ ($2^\tau \gg n$) has not shown any negative effects in our preliminary experiments. Using bit vectors comes with the advantage  that genetic operators like mutation and recombination are trivially defined on this kind of data structure.

Formally, to any individual $x \in X$ we assign a bit vector $T(x) = \langle t_0, ..., t_{\tau-1} \rangle$ with $t_i \in \{0, 1\}$ for all $i$, which is initialized at random when the individual $x$ is created. Every time a mutation operation is performed on $x$, we perform a random single bit flip on $T(x)$.\footnote{Note that this works for typical mutation operators on the non-trash genes, which tend to change very little about the genome as well. If more invasive mutation operators are employed, a likewise operation on the bit vector could be provided.} For each recombination of $x_1$ and $x_2$, we generate the child's trash bit vector via uniform crossover of $T(x_1)$ and $T(x_2)$.

We can then compute a trash bit distance $\textit{tdist}$ between two individuals $x_1$ and $x_2$ simply by returning the Hamming distance between their respective trash genes:

\begin{align*} 
\textit{tdist}(x_1, x_2) & = \frac{1}{\tau} \; * \; H(T(x_1), T(x_2)) \\ & = \frac{1}{\tau} \; * \; \sum_{i=0}^{\tau - 1} |T(x_1)_i - T(x_2)_i|
\end{align*}

This metric clearly is symmetric. Again, we normalize the output by dividing it by $\tau$. Furthermore, trash bit vectors allow for a more detailed distinction between the impact of various genetic operators: The expected distance between two randomly generated individuals $x_1$ and $x_2$ is $\mathbb{E}(\textit{tdist}(x_1, x_2)) = 0.5$. However, the distance between parents and children is reasonably lower: If $x$ is the result of mutating $x'$, we expect their trash bit distance to be $\mathbb{E}(\textit{tdist}(x, x')) = 1/\tau$. The trash bit distance between a parent $x'$ of a recombination operator and its child $x$ is $\mathbb{E}(\textit{tdist}(x, x')) = 0.25$ since the child shares about half of its trash bits with this one parent $x'$ by the nature of crossover, resulting in a Hamming distance of $0$ on this subset, and the other half with the other parent, say $x''$, with which the first parent $x'$ naturally shares about half of its trash bits when no other assumptions about the parents' ancestry apply. This means that for the subset of trash bits inherited from $x''$, $x$ and $x'$ have a trash bit distance of $0.5$, resulting in a $0.25$ average for the whole bit vector of $x$.\footnote{These numbers correspond closely to the degrees of genetic relationship mentioned in \cite{dawkins2016selfish}.} If the parents are related (or share improbable amounts of trash bits by chance), lower numbers for $\textit{tdist}$ can be achieved.

These examples should illustrate that the computed trash bit diversity is able to express genealogical relations between individuals. It stresses recombination over mutation but in doing so reflects the impact the respective operators have on the individual's actual genome. We thus propose trash bit vectors as a much simpler and more efficient implementation of genealogical diversity.

As is clear from the usage of the ``expected value'' $\mathbb{E}$ in these computations, the actual distance between parents and offspring is now always subject to random effects. However, so is their similarity on the non-trash genes as well.\footnote{For example, a child generated via uniform crossover has very slim chance of not inheriting any gene material from one parent at all. The same effect can now happen not only on the fitness-relevant genes but also on the genes used for diversity marking.} This kind of probabilistic behavior is an intrinsic part of evolutionary algorithms. However, it may make sense to base the recombination on the trash bit vector on the recombination of the non-trash genes so that probabilistic tendencies are kept in sync. This is up to future research.

Finally, the computational effort to compute the trash bit distance is at most times negligible. Computing the distance between two individuals is an operation that can be performed in $O(\tau)$ and while we expect there to be a lose connection between population size and the optimal $\tau$, for a given evolution process with a fixed population size, this means that trash bit diversity can be computed in constant time. Trivially, this also means the concept scales with population size and age.

\section{Experiment}
\label{experiment}

\begin{figure}[t]
  \centering
  \includegraphics[width = 0.45\textwidth]{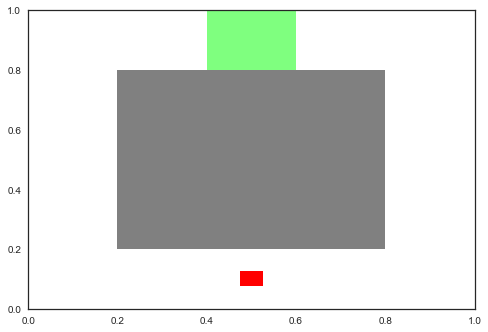}
  \caption{Illustration of the setup of the scenario. Marked in red is the starting position of the agent. The green area defines the goal which the agent is supposed to drive to. The gray area is the main obstacle the robot needs to drive around.}
  \label{fig:setup}
\end{figure}

To verify the practical applicability of the concept of genealogical diversity and its realizations presented in the previous Sections~\ref{tree} and \ref{trash}, respectively, we constructed a simple experimental setup: We define a simple routing task in which a robot has to choose a sequence of $10$ continuous actions $a \in \mathbb{R} \times \mathbb{R}$ to reach a marked area. Each action takes exactly one time step and can move the robot across a Manhattan square of $0.5$ at most. For each time step the robot remains inside the designated target area, it is rewarded a bonus of $+1$.   In order to reach that area, the robot has to find a path around an obstacle blocking the direct way. Figure~\ref{fig:setup} shows a simple visualization of the setup described here.

\begin{figure}[t]
  \centering
  \includegraphics[width = 0.45\textwidth]{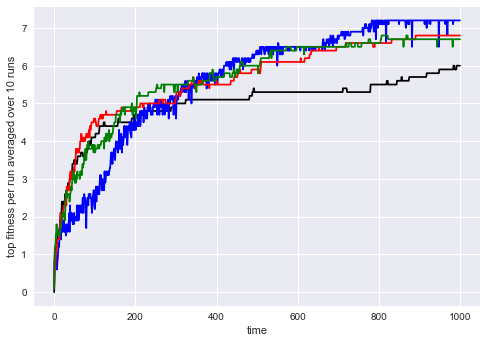}
  \caption{Average fitness achieved over time by the the non-diverse genetic algorithm (black), the domain-specific diverse genetic algorithm (blue), the genealogically diverse algorithm based on the genealogical tree distance (green) and the genealogically diverse genetic algorithm using the trash bit distance (red). To mitigate random effects a bit, the fitness values have been averaged over 10 complete evolution runs.}
  \label{fig:plot}
\end{figure}

We solved this scenario with four different evolutionary algorithms. All of these use a population size of $20$ individuals and have been executed for $1000$ generations. For this kind of continuous optimization problem, that is not enough time for them to fully converge. We constructed a standard setup of an evolutionary algorithm with a continuous mutation operator working on a single action at a time and activated with a probability of $0.2$. We employ uniform crossover with a probability of $0.3$ per individual. A recombination partner is selected from a two-player tournament and offspring is added to the population before the selection step. Furthermore, $2$ new individuals per generation are generated randomly.

Within this setup, we define a standard genetic algorithm using a fitness function that simply returns the aforementioned bonus for each individual. It performs well but seems to suffer from premature convergence in this setup (see Figure~\ref{fig:plot} for all plots). This is the baseline approach all diversity-enabled versions of the genetic algorithm can be tested against.

To introduce the diversity of the solutions to the genetic algorithm, we choose the approach to explicitly include the distance of the individual $x$ to other individuals of $P$ in $x$'s fitness. But we do not construct a multi-objective optimization problem (as in \cite{laumanns2002combining, segura2016using}, e.g.) but simply define a weighting function to flatten these objectives. Formally, given the original fitness function $f$ and an average diversity measure $d$ of a single individual with respect to the population $P \subseteq X$, we define an adapted fitness function $f'$ as follows:

$$f'(x, P) = f(x) + \lambda * d(x, P)$$

It is important to note that while we use $f'$ for the purpose of selection inside the evolutionary algorithm, all external analysis (plotting, e.g.) is performed on the value of $f$ only in order to keep the results comparable. Also note that we reduce the computational effort to calculate any distance metric $d$ used in this paper by not evaluating a given individual's diversity against the whole population $P$ but only against a randomly chosen subset of $5$ individuals. In our experiments, this approach has been sufficiently stable.

Furthermore, we determined the optimal $\lambda$ for each algorithm using grid search on this hyperparameter. In a scenario like this, where higher diversity yields better results overall, it appears reasonable to think that $\lambda$ could be determined adaptively during the evolutionary process. This is still up to further research.

For evaluation purposes, we provided a domain-specific distance function. In this simple scenario, this can be defined quickly as well and we chose to use the sum of all differences between actions at the same position in the sequence. Figure~\ref{fig:plot} shows that this approach takes a bit longer to learn but can then evade local optima better, showing a curve that we would expect from a more diverse genetic algorithm.

Finally, we implemented both genealogical distance metrics presented in this paper. We can see in Figure~\ref{fig:plot} that both approaches in fact perform comparably, even though trash bit vectors require much less computational effort. For this experiment, we used $\tau = 32$.


\section{Conclusion}
\label{conclusion}

In this paper, we have introduced the expected evolutionary edit distance as a promising target for diversity-aware optimization within evolutionary algorithms. Having found that it cannot be reasonably computed within another evolutionary process, we developed approaches to estimate that distance more efficiently. To do so, we introduced the notion of genealogical diversity and presented a method to estimate it accurately using very little computational resources.

The experimental results show the initial viability of the approach used here and allow for many future applications. Some of these have been realized in \cite{gabor2017icac}. Other promising directions for future work have been mentioned throughout and include plans to omit the hyperparameter $\lambda$ by using genealogical diversity within a true multi-objective evolutionary algorithm or by opening $\lambda$ for self-adaptation by the evolutionary process. Furthermore, from a biological point of view, a genealogical selection process is less common in survivor selection than it is in parent selection. Testing if the metaphor to biology holds in that case would be an immediate next step of research.

%

\bibliographystyle{ACM-Reference-Format}
\bibliography{references} 

\end{document}